\pdfoutput=1

\documentclass[11pt]{article}

\usepackage{acl}

\usepackage{times}
\usepackage{latexsym}

\usepackage[T1]{fontenc}

\usepackage[utf8]{inputenc}

\usepackage{microtype}

\usepackage{url}

\usepackage{booktabs}
\usepackage{multirow}
\usepackage{subcaption}

\usepackage{amssymb} 
\usepackage{amsmath}

\usepackage{CJKutf8}

\usepackage{graphbox}

\usepackage{tikz}
\usepackage{tikz-qtree}
\usetikzlibrary{arrows,decorations.pathmorphing,backgrounds,positioning,fit,petri,shapes.misc, arrows.meta,shapes.geometric,decorations.markings,calc,shadows.blur,decorations.pathreplacing,quotes}
\definecolor{myblue}{RGB}{6, 82, 221}
\definecolor{myorange}{RGB}{211, 84, 0}
\definecolor{lowblue}{RGB}{102,178,255}
\definecolor{justblue}{RGB}{84, 160, 255}
\definecolor{mypurple}{RGB}{108, 92, 231}
\definecolor{mygray}{RGB}{158, 158, 158}
\definecolor{lowpurple}{RGB}{204,153,255}
\definecolor{lowwhite}{RGB}{255,255,255}
\definecolor{verylowpurple}{RGB}{255,102,102}
\definecolor{embcolor}{RGB}{255,255,255}
\definecolor{myred}{RGB}{235, 47, 6} 
\definecolor{mygreen}{RGB}{162, 217, 206} 
\definecolor{fontgrey}{RGB}{44, 62, 80}
\definecolor{lowpurple}{RGB}{210, 180, 222}
\definecolor{mypumpkin}{RGB}{229, 152, 102}
\definecolor{lowgreen}{RGB}{171, 235, 198}
\definecolor{lowgreen2}{RGB}{186, 220, 88}
\definecolor{lowred}{RGB}{245, 183, 177}
\definecolor{lowyellow}{RGB}{241, 196, 15}
\definecolor{mypink}{RGB}{255, 118, 117}
\definecolor{bluemartina}{RGB}{18, 203, 196}
\definecolor{puffin}{RGB}{250, 152, 58}
\definecolor{grass}{RGB}{0, 148, 50}
\definecolor{cnngray}{RGB}{116, 125, 140}

\newcommand{\squishlist}{
	\begin{list}{$\bullet$}
		{ \setlength{\itemsep}{0pt}
			\setlength{\parsep}{3pt}
			\setlength{\topsep}{3pt}
			\setlength{\partopsep}{0pt}
			\setlength{\leftmargin}{1.5em}
			\setlength{\labelwidth}{1em}
			\setlength{\labelsep}{0.5em} } }

	\newcounter{Lcount}
	\newcommand{\squishlisttwo}{
		\begin{list}{\arabic{Lcount}. }
			{ \usecounter{Lcount}
				\setlength{\itemsep}{0pt}
				\setlength{\parsep}{0pt}
				\setlength{\topsep}{0pt}
				\setlength{\partopsep}{0pt}
				\setlength{\leftmargin}{2em}
				\setlength{\labelwidth}{1.5em}
				\setlength{\labelsep}{0.5em} } }
		
		\newcommand{\squishend}{
	\end{list} }

\usepackage{multirow}
\usepackage{hhline}
\usepackage{tabularx}
\usepackage{ragged2e}
\newcolumntype{Y}{>{\RaggedRight\let\newline\\\arraybackslash\hspace{0pt}}X} 
\usepackage{xcolor}
\usepackage{pgfplots, pgfplotstable}
\usepackage{pgfpages}
\usepackage{mathbbol}
\usepackage{arydshln}
\usepackage{graphicx}

\usepackage{amssymb}
\usepackage{amsfonts}

\usepackage{algorithm}
\usepackage{algorithmic}
\usepackage{pgf}

\usepackage{lipsum}
\usepackage{multicol}
\usepackage{changepage}

\usepackage[export]{adjustbox}

\usepackage[utf8]{inputenc}
 
\usepackage{amsthm}

\usepackage{flushend}

\title{IAM: A Comprehensive and Large-Scale Dataset for Integrated Argument Mining Tasks}

\author{
\textbf{
Liying Cheng\thanks{~~Liying Cheng is under the Joint Ph.D. Program between Alibaba and Singapore University of Technology and Design.} \textsuperscript{\rm ~1,2}~~~ 
Lidong Bing\thanks{$^\dag$ Corresponding author.}$^\dag$\textsuperscript{\rm 1}~~~
Ruidan He\textsuperscript{\rm 1}~~~
Qian Yu\thanks{$^\ddag$Qian Yu and Yan Zhang were interns at Alibaba.}$^\ddag$\textsuperscript{\rm 3}~~~
Yan Zhang$^\ddag$\textsuperscript{\rm 4}~~~
Luo Si\textsuperscript{\rm 1}}\\
\textsuperscript{\rm 1}DAMO Academy, Alibaba Group~~
\textsuperscript{\rm 2}Singapore University of Technology and Design
\\
\textsuperscript{\rm 3}The Chinese University of Hong Kong ~~
\textsuperscript{\rm 4} National University of Singapore\\
{\tt\{liying.cheng, l.bing, ruidan.he, luo.si\}@alibaba-inc.com}
\\
{\tt yuqian@se.cuhk.edu.hk~~~eleyanz@nus.edu.sg}
}

\begin{document}
\maketitle
\begin{abstract}
Traditionally, a debate usually requires a manual preparation process, including reading plenty of articles, selecting the claims, identifying the stances of the claims, seeking the evidence for the claims, etc.
As the AI debate attracts more attention these years, it is worth exploring the methods to automate the tedious process involved in the debating system.
In this work, we introduce a comprehensive and large dataset named IAM, which can be applied to a series of argument mining tasks, including claim extraction, stance classification, evidence extraction, etc.
Our dataset is collected from over 1k articles related to 123 topics.
Near 70k sentences in the dataset are fully annotated based on their argument properties (e.g., claims, stances, evidence, etc.).
We further propose two new integrated argument mining tasks associated with the debate preparation process: (1) claim extraction with stance classification (CESC) and (2) claim-evidence pair extraction (CEPE).
We adopt a pipeline approach and an end-to-end method for each integrated task separately.
Promising experimental results are reported to show the values and challenges of our proposed tasks, and motivate future research on argument mining.
\footnote{Our code, data and leaderboard are available at \url{https://github.com/LiyingCheng95/IAM}.}

\end{abstract}

\section{Introduction}
Debating has a long history and wide application scenarios in education field \cite{stab2014identifying,persing2016end,stab2017parsing}, political domain \cite{lippi2016argument,duthie2016mining,menini2018never}, legal actions \cite{mochales2011argumentation,grabmair2015introducing,teruel2018increasing}, etc.
It usually involves tons of manual preparation steps, including reading the articles, selecting the claims, identifying the claim stances to the topics, looking for the evidence of the claims, etc.
Since the machine has shown promising potential in processing large quantities of information in many other natural language processing tasks, it is also worthwhile to explore the methods for automating the manual process involved in debating. 

\begin{figure}[t!]
    \center{
    \includegraphics[width=\linewidth]
    {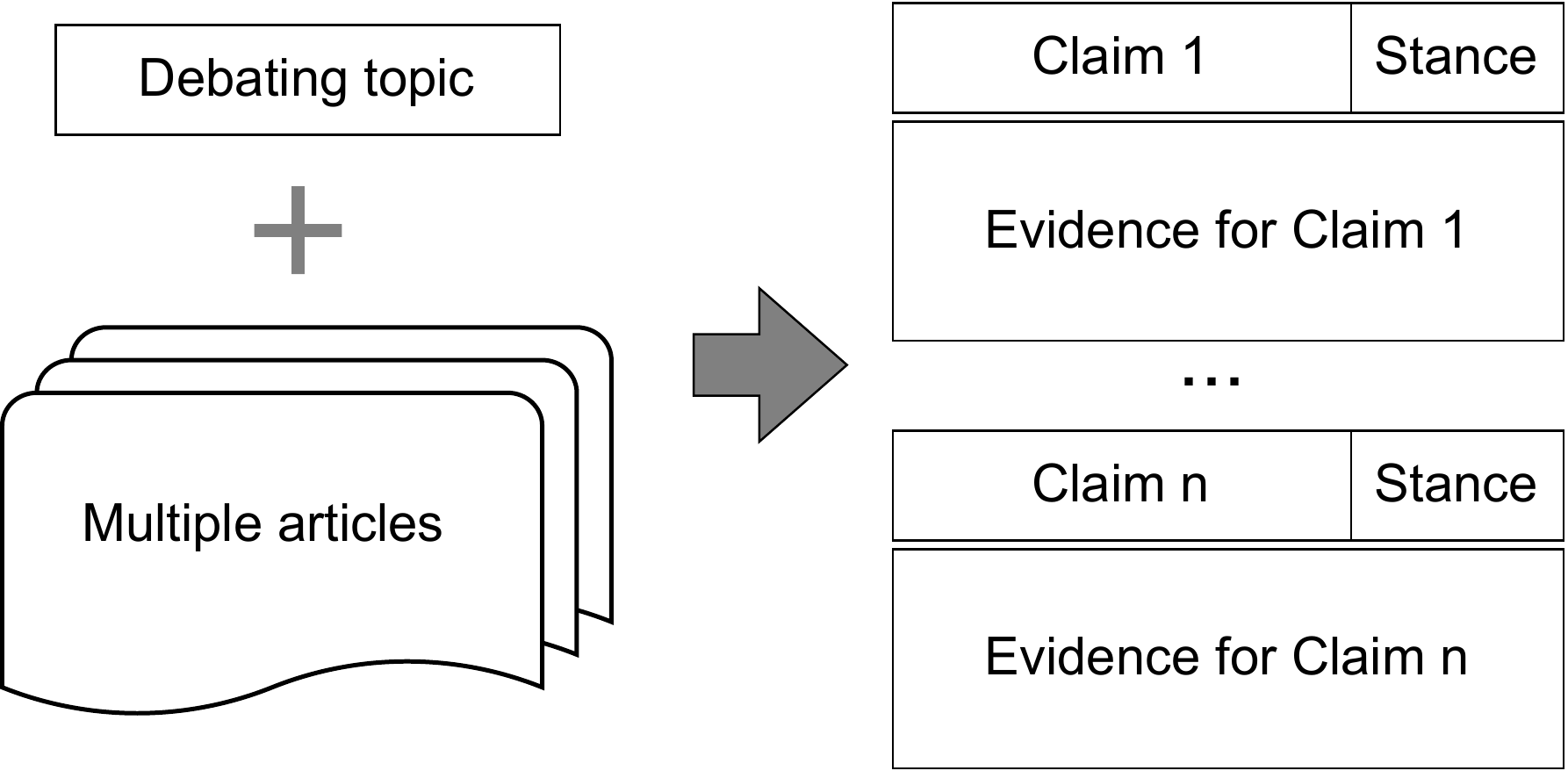}}
    \caption{\label{fig:flow} A flow chart showing the debating preparation process. }
\end{figure}

Argument mining (AM), as the core of a debating system \cite{bar2021advances, yuan2021overview}, has received more attention in the past few years.
Several AM tasks and datasets have been proposed to work towards automatic AI debate, such as: context dependent claim detection (CDCD) \cite{levy2014context}, claim stance classification (CSC) \cite{bar2017stance, chen2019seeing}
, context dependent evidence detection (CDED) \cite{rinott2015show}, etc.
All the above tasks are essential elements for AM and they are mutually reinforcing in the debating preparation process.
In this work, we aim at automating the debating preparation process as shown in Figure \ref{fig:flow}. 
Specifically, providing with the debating topic and several related articles, we intend to extract the claims with their stances, and also the evidence supporting the claims.

However, none of the existing works can facilitate the study of all these tasks at the same time.
Motivated by this, we introduce a comprehensive dataset named IAM to support the research of these tasks.
We create our dataset by first collecting over 100 topics from online forums and then exploring over 1k articles related to these topics. 
All the sentences in those articles are fully-annotated following a set of carefully defined annotation guidelines.
Given a specific topic, the annotators have to distinguish whether the given sentence is a claim to this topic and identify the relation between the selected claim and the topic (i.e., support or contest).
Then given the claims, the annotators have to browse the contexts to find evidence supporting the claims.
With all the labeled information, researchers can work towards these primary argument mining tasks simultaneously.

To better coordinate these individual tasks together, we propose two new integrated tasks: claim extraction with stance classification (CESC) and claim-evidence pair extraction (CEPE).
Instead of treating the existing tasks (i.e., CDCD, 
CSC,
CDED) as individual ones, the two proposed tasks can integrate the relevant primary tasks together, which are more practical and more effective in the debating preparation process.
CESC can be divided into two subtasks: the claim detection task and the stance classification task.
Intuitively, we conduct experiments on CESC with a pipeline approach to combine the two subtasks.
As the two subtasks are mutually reinforcing each other, we also adopt an end-to-end classification model with multiple labels (i.e., support, contest, and no relation).
CEPE is composed of the claim detection task and the evidence detection task.
Similar to the annotation procedure, we apply a pipeline method to tackle this problem by first detecting the claims given the topics and then identifying the corresponding evidence of each claim.
We also use a multi-task model to extract both claims and evidence as well as their pairing relation simultaneously.
We conduct extensive experiments on our dataset to verify the effectiveness of our models and shed light on the challenges of our proposed tasks.

To summarize, our contributions are as follows.
(1) We introduce a fully-annotated argument mining dataset and provide thorough data analysis. This is the first dataset that supports comprehensive argument mining tasks. 
(2) We are the first to propose the CESC and CEPE tasks, which are practical task settings in the argument mining field and able to enlighten future research on this.
(3) We conduct preliminary experiments for all proposed tasks with the new dataset. 

\section{Related Work}

In recent years, there is a tremendous amount of research effort in the computational argumentation research field \cite{eger2017neural,bar2021advances}, such as argument components identification \cite{levy2014context,rinott2015show,lippi2016argument,daxenberger2017essence}, argument classification and clustering \cite{reimers2019classification}, argument relation prediction \cite{boltuvzic2016fill, chakrabarty2019ampersand}, argument pair extraction \cite{cabrio2012combining, cheng2020argument, cheng2021argument, yuan2021leveraging, ji2021discrete}, argument quality assessment \cite{habernal2016argument,wachsmuth2017computational,gretz2020large,toledo2019automatic}, listening comprehension \cite{mirkin2018listening}, etc.

Meanwhile, researchers have been exploring new datasets and methods to automate the debating preparation process, such as project debater \cite{slonim2021autonomous}, etc.
\citet{bilu2019argument} work on the argument invention task in the debating field to automatically identify which of these arguments are relevant to the topic.
\citet{li2020exploring} explore the role of argument structure in online debate persuasion.
\citet{levy2014context} introduce a dataset with labeled claims and work on the task of context-dependent claim detection (CDCD). 
\citet{bar2017stance} modify \citet{aharoni2014benchmark}'s dataset by further labeling the claim stances, and tackle the problem of stance classification of context-dependent claims.
\citet{rinott2015show} propose a task of detecting context-dependent evidence that supports a given claim (CDED) and also introduce a new dataset for this task.

Unlike previous works with a specific focus on only one argument mining task, we introduce a comprehensive dataset that is able to support different tasks related to the debating system.
Such a dataset not only enlightens future research on the argument mining field but also shows strong potential for various practical applications.
Another difference is that existing tasks (e.g., CDCD, CDED, CSC, etc.) could be considered as subtasks in the emerging wider field of argumentation mining \cite{levy2014context}. 
While in this paper, we propose two integrated tasks (i.e., CESC and CEPE) incorporating the existing subtasks in the debating system, which takes a step forward to automatic AI debate.
A more detailed comparison to the most representative and relevant previous datasets will be shown in Section \ref{sec:data_analysis}.







\begin{table*}[t!]
	\centering
	\resizebox{\linewidth}{!}{
        \begin{tabular}{c p{13cm} ccc}
        \toprule
         & \textbf{Topic: Will artificial intelligence replace humans} & \textbf{Claim} & \textbf{Stance} & \textbf{Evidence} \\
        \toprule
\multirow{2}{*}{1} & Job opportunities will grow with the advent of AI; however, some jobs might be lost because AI would replace them. & \multirow{2}{*}{C\_1} & \multirow{2}{*}{+1} & \\\cdashline{1-5}
2 & Any job that involves repetitive tasks is at risk of being replaced. & C\_2 & +1 & \\\cdashline{1-5}
\multirow{2}{*}{3} & In 2017, Gartner predicted 500,000 jobs would be created because of AI, but also predicted that up to 900,000 jobs could be lost because of it. & & & \multirow{2}{*}{E\_1 | E\_2} \\\midrule
4 & The number of industrial robots has increased significantly since the 2000s. & & & E\_3 \\\cdashline{1-5}
5 & The low operating costs of robots make them competitive with human workers. & & & E\_3 \\\cdashline{1-5}
\multirow{2}{*}{6} & In the finance sector, computer algorithms can execute stock trades much faster than a human, needing only a fraction of a second. & & & \multirow{2}{*}{E\_3} \\\cdashline{1-5}
\multirow{2}{*}{7} & As these technologies become cheaper and more accessible, they will be implemented more widely, and humans might be increasingly replaced by AI. & \multirow{2}{*}{C\_3} & \multirow{2}{*}{+1} & \\\midrule
\multirow{2}{*}{8} & According to Harvard Business Review, most operations groups adopting RPA have promised their employees that automation would not result in layoffs. & \multirow{2}{*}{C\_4} & \multirow{2}{*}{-1} & \multirow{2}{*}{E\_4} \\\midrule
9 & AI is incredibly smart, but it will never match human creativity. & C\_5 & -1 & \\
        \bottomrule
        \end{tabular}}
    \caption{Sample topic and labeled claims with their stances and evidence. Note that different blocks refer to the sentences from different articles, and we only extract claim-evidence pairs from the same article.
    For clarity, we label the indices in ascending order, which may not reflect the real indices in the dataset.
    }
	\label{tab:data1}
\end{table*}

\section{IAM Dataset}

We introduce a large and comprehensive dataset to facilitate the study of several essential AM tasks in the debating system.
We describe the collection process, annotation details and data analysis here.

\subsection{Data Collection}

First, we collect 123 debating topics with a wide variety from online forums. 
For each topic, we explore around 10 articles from English Wikipedia with promising content.
The most number of articles explored for one topic is 16, while the least number is 2.
This is because it is difficult to find enough resources for unpopular topics such as ``Should nuclear waste be buried in the ground''.
However, most topics (i.e., 91 topics) are relatively popular with more than 8 related articles collected for each of them.
In total, there are 1,010 articles collected for all the topics.
After we obtain all the relevant articles, we use the NLTK package \cite{bird2009natural} to split the corpus into 69,666 sentences from these articles for further annotation.

\begin{table*}[t!]
	\centering
	\resizebox{\linewidth}{!}{
        \begin{tabular}{lc|cc|cccc|c|c}
        \toprule
        & \multirow{2}{*}{Topics} & \multirow{2}{*}{Articles} & Articles & \multirow{2}{*}{Claims} & \multirow{2}{*}{Support} & \multirow{2}{*}{Contest} & Claims with & \multirow{2}{*}{Evidence} & \multirow{2}{*}{CEPs} \\
        & & & with claims & & & & evidence & & \\
        \midrule
        \citet{levy2014context} & 32 & 326 & - & 976 & - & - & - & - & - \\
        \citet{rinott2015show} & 39 & 274 & 274 & 1,734 & - & - & 1,040 & 3,057 & 5,029* \\
        \citet{aharoni2014benchmark} & 33 (12) & 586 & 321 (104) & 1,392 & - & - & (350) & (1,291) & 1,476* \\
        \citet{bar2017stance} & 55 & - & - & 2,394 & 1,324 & 1,070 & - & - & - \\
        IAM (Ours) & 123 & 1,010 & 814 & 4,890 & 2,613 & 2,277 & 3,302 & 9,384 & 10,635 \\
        \bottomrule
        \end{tabular}}
    \caption{Overall statistics comparison of the existing datasets and our dataset. 
    Note that: (1) in \citet{aharoni2014benchmark}'s dataset, the numbers in the parenthesis refer to the evidence labeling data; (2) the numbers with * are calculated by us since they are not shown in the original papers.}
	\label{tab:data2}
\end{table*}

\begin{table}[t!]
	\centering
	\resizebox{\columnwidth}{!}{
	    \setlength{\tabcolsep}{4mm}{
        \begin{tabular}{lc}
        \toprule
        & Ours \\
        \midrule
        Avg. length of all sentences & 21.05 \\
        Avg. length of claims & 23.44 \\
        Avg. length of evidence & 25.09 \\
        \midrule
        Avg. \% vocab shared in each CEP & 20.14\% \\
        Avg. \% vocab shared in each sent pair & 8.73\% \\
        \bottomrule
        \end{tabular}}
        }
    \caption{Dataset statistics on argument lengths and vocabulary sharing.}
	\label{tab:data3}
\end{table}

\subsection{Data Annotation}
\label{sec:data_annotation}
The annotation process is mainly separated into two stages: (1) detecting the claims given the topics, (2) detecting the evidence given the claims.
A context-dependent claim (CDC), claim in short, is a general and concise statement that directly supports or contests the given topic \cite{levy2014context}.
The annotators are asked to extract the claims by following this definition. 
Meanwhile, the annotators have to identify the stance of the extracted claim towards the given topic.
In the second stage, the annotators have to read through the context surrounding the claims, and extract the evidence following that a piece of context-dependent evidence (CDE) is a text segment that directly supports a claim in the context of the topic.
Since only the surrounding sentences are content-relevant in most cases, we only search 10 to 15 sentences before and after the claim sentence to label the evidence.
Note that the claim itself could be the evidence as well.

Professional data annotators are hired from a data annotation company and are fully paid for their work.
Each sentence is labeled by 2 professional annotators working independently in the first round.
69,666 sentences are labeled in total and the Cohen's kappa is 0.44 between the two annotators, which is a reasonable and relatively high agreement considering the annotation complexity \cite{aharoni2014benchmark, levy2014context}.
Whenever there is any inconsistency, the third professional annotator will judge the annotation result in the confirmation phase to resolve the disagreement.

Table \ref{tab:data1} shows a sample topic ``Will artificial intelligence replace humans'' and its labeled claims with their stances and evidence.
The claims are labeled as ``C\_index'' and the evidence is labeled as ``E\_index''.
For stances, ``+1'' represents the current claim supporting the topic, while ``-1'' represents the claim contesting the topic.
A claim and a piece of evidence form a claim-evidence pair (CEP) if the indices match with each other under a specific topic.
A piece of evidence can support multiple claims, such as Sent 3, as a piece of evidence, it supports two claims, i.e. Sent 1 and 2.
Similarly, a claim can have different evidence, such as Sent 7, as a claim, it has three paired evidence sentences (i.e., Sent 4 - 6).
As mentioned, one sentence can be considered as both the claim and the evidence.
For instance, in Sent 8, there is a clear and concise statement ``automation would not result in layoffs'' contesting the given topic directly, which is considered as a claim.
There is also a text segment at the beginning of the sentence showing the testimony from an organization (i.e., ``Harvard Business Review'') directly supporting this claim stated in the latter part of the sentence. Therefore, this sentence is labeled as evidence as well.
Last but not least, there are some claims without evidence found in the context in our dataset, such as Sent 9.

\subsection{Dataset Analysis}
\label{sec:data_analysis}

We present the dataset statistics comparison with existing datasets in Table \ref{tab:data2}, and list the key differences below.
First, as mentioned earlier, the existing datasets have their own focus on particular tasks, and none of them can support all the essential argument mining tasks related to the debate preparation process.
\citet{levy2014context} only label data for claims, \citet{rinott2015show} only focus on detecting the evidence given the claims, \citet{aharoni2014benchmark} only label a partial dataset for evidence, and \citet{bar2017stance} only tackle the claim stance classification problem.
In contrast, our dataset is fully annotated for all the key elements related to argument mining tasks, including claims, stances, evidence, and relations among them.
Although combining \citet{aharoni2014benchmark} and \citet{bar2017stance}'s datasets can obtain a comprehensive dataset with 12 topics supporting all the subtasks,
in terms of the dataset size, our dataset is significantly larger than it and the existing datasets.
We explore 123 topics in total, which is more than twice of \citet{bar2017stance}'s dataset.
Accordingly, we obtain much more claims and evidence by human annotation on all sentences in the corpus, as compared to the previous datasets, which could add potential value to the argument mining community.

Table \ref{tab:data3} shows more statistics of our dataset.
In terms of the sentence lengths in our dataset, the average number of words in a sentence is around 21.
The average length of sentences containing claims is generally longer, and evidence is even slightly longer.
However, since the length differences are subtle, it shows the challenges to distinguish the claims and evidence using the length differences among the sentences.
We also calculate the average percentage of vocabulary shared between each claim-evidence sentence pair, which is 20.14\%; while the same percentage between any two sentences from our corpus is only 8.73\%.
This shows that extracting CEP is a reasonable task as it has a higher percentage of vocabulary sharing than other sentence pairs, but it is also challenging as the absolute percentage is still low.


\section{Tasks}

In the debating system, our ultimate goal is to automate the whole debate preparation process as shown in Figure \ref{fig:flow}. With the introduced annotated dataset, we can tackle all core subtasks involved in the process at the same time. In this section, we first review the existing subtasks, and then propose two integrated argument mining tasks.

\subsection{Existing Tasks}
\paragraph{Task 1: Claim Extraction}
Similar to the CDCD task proposed by \citet{levy2014context}, this task is defined as: given a specific debating topic and related articles, automatically extract the claims from the articles.
Claim extraction is a primary argument mining task as the claim is a key argument component.

\paragraph{Task 2: Stance Classification}
As introduced by \citet{bar2017stance}, this task is defined as: given a topic and a set of claims extracted for it, determine for each claim whether it supports or contests the topic.
As shown in Table \ref{tab:data2}, the number of claims from two stances is approximately balanced (i.e., 53.4\% are support and 46.6\% are contest).

\paragraph{Task 3: Evidence Extraction}
In \citet{rinott2015show}'s work, this task is defined as: given a concrete topic, a relevant claim, and potentially relevant documents, the model is required to automatically pinpoint the evidence within these documents. In this paper, we only explore the evidence candidate sentences from the surrounding sentences of the claims, as long-distance sentences may not be content-relevant in most cases.

\subsection{Integrated Tasks}

In order to further automate the debating preparation process, exploring integrated tasks rather than individual subtasks is non-trivial. In this work, we introduce two integrated argument mining tasks as below to better study the subtasks together.

\paragraph{Task 4: Claim Extraction with Stance Classification (CESC)}
Since claims stand at a clear position towards a given topic, the sentences with clear stances should have a higher possibility to be the claims.
Hence, identifying the stances of the claims is supposed to benefit the claim extraction task.
By combining Task 1 and Task 2, we define the first integrated task as: given a specific topic and relevant articles, extract the claims from the articles and also identify the stance of the claims towards the topic.

\paragraph{Task 5: Claim-Evidence Pair Extraction (CEPE)}
Since evidence is clearly supporting the corresponding claims in an article, claims and evidence are mutually reinforcing each other in the context. Therefore, we hypothesize the claim extraction task and the evidence extraction task may benefit each other.
By combining Task 1 and Task 3, we define the second integrated task as: given a specific topic and relevant articles, extract the claim-evidence pairs (CEPs) from the articles.

\section{Approaches}

To tackle the two integrated tasks, we first adopt a pipeline approach to pipe the corresponding subtasks together by using sentence-pair classification on each subtask. We also propose two end-to-end models for the two integrated tasks.

\subsection{Sentence-pair Classification}
\label{sec:sentpairclassification}
We formulate Task 1, Task 2, and Task 3 as sentence-pair classification tasks.
We train a sentence-pair classifier based on pre-trained models such as BERT \cite{devlin2019bert} and RoBERTa \cite{liu2019roberta}.
The sentence pairs are concatenated and fed into the pre-trained model to get the hidden state of the ``[CLS]'' token.
Then, a linear classifier will predict the relation between the two sentences.
Specifically, for Task 1, the topic and the article sentence are concatenated and fed into the model.
If they belong to the same pair, the article sentence is considered as a claim, and vice versa.
For Task 2, the model predicts the stance between a topic and a claim.
Task 3 is similar to Task 1, where the model predicts if the given claim and the article sentence form a pair, i.e., if the sentence is a piece of evidence of the claim.
All these three tasks can be considered as binary classification tasks, and cross-entropy loss is used as the loss function.

\paragraph{Negative Sampling}
For Task 1 and Task 3, the binary labels are unbalanced as the number of claims and pieces of evidence is far smaller than the total number of sentences. 
To overcome this difficulty, we adopt negative sampling techniques \cite{mikolov2013distributed}. 
During the training of these two tasks, for each claim/evidence sentence, we randomly select a certain amount of non-claim/non-evidence sentences as negative samples. 
These negative samples together with all claims/evidence form a new training dataset for each task.

\subsection{Multi-Label Model for CESC}
Apart from the pipeline approach, we propose a multi-label model for CESC.
Instead of handling the two subtasks separately, we concatenate the topic and article sentences to feed into a pre-trained model and define 3 output labels specifically for this task: support, contest, and no-relation.
Support and contest refer to those claims with their corresponding stances to the topic, while no-relation stands for non-claims.
Since the sentence pairs with no-relation labels are much more than those with support/contest, we also apply negative sampling here for a more balanced training process.

\subsection{Multi-Task Model for CEPE}
Inspired from \citet{cheng2021argument}'s work, we adopt a multi-task model (i.e., an attention-guided multi-cross encoding-based model) for the CEPE task.
Provided with a sequence of article sentences and the topic, we first concatenate the topic and individual sentences as the claim candidates, and use the sequence of article sentences as the evidence candidates.
We reformulate the claim extraction and evidence extraction subtasks as sequence labeling problems.
Then, the sequence of claim candidates and the sequence of evidence candidates go through the pre-trained models to obtain their sentence embeddings respectively.
To predict whether two sentences form a claim-evidence pair, we adopt a table-filling approach by pairing each sentence in the claim candidates with each sentence in the evidence candidates to form a table.
All three features (i.e., claim candidates, evidence candidates, table) update each other through the attention-guided multi-cross encoding layer as described in \citet{cheng2021argument}'s work.
Lastly, the two sequence features are used to predict their sequence labels, the table features are used for pair prediction between each claim and evidence.
Compared to the pipeline approach, this multi-task model has stronger subtask coordination capability, as the shared information between the two subtasks is learned explicitly through the multi-cross encoder.

\section{Experiments}

\subsection{Experimental Settings}
\begin{table}[t!]
	\centering
	\resizebox{\columnwidth}{!}{
        \begin{tabular}{lccc}
        \toprule
        & train & dev & test \\
        \midrule
        \# sents as claim candidates & 55,544 & 7,057 & 7,065 \\
        \# claims & \textcolor{white}{0}3,871 & \textcolor{white}{0,}492 & \textcolor{white}{0,}527\\
        ~~~~\# support claims & \textcolor{white}{0}2,098 & \textcolor{white}{0,}259 & \textcolor{white}{0,}256 \\
        ~~~~\# contest claims & \textcolor{white}{0}1,773 & \textcolor{white}{0,}233 & \textcolor{white}{0,}271 \\
        \# claims with evidence & \textcolor{white}{0}2,616 & \textcolor{white}{0,}347 & \textcolor{white}{0,}375 \\
        ~~~~\% claims with evidence & \textcolor{white}{0}67.6\% & 70.3\% & 71.2\% \\
        \# sents as evidence candidates & 57,398 & 7,487 & 8,172 \\
        \# pieces of evidence & \textcolor{white}{0}7,278 & \textcolor{white}{0,}909 & 1,108 \\
        Avg. \# pieces of evidence per claim & \textcolor{white}{00}2.78 & \textcolor{white}{0}2.62 & \textcolor{white}{0}2.95 \\
        \bottomrule
        \end{tabular}}
    \caption{Dataset statistics split on train/dev/test sets.}
	\label{tab:data4}
\end{table}

We split our dataset randomly by a ratio of 8:1:1 for training, development, and testing.
The dataset statistics are shown in Table \ref{tab:data4}.
In the training set, since the number of claims (3,871) and the number of non-claims (51,673) are not balanced with a ratio of 1:13.3, we conduct experiments by selecting different numbers of negative samples and evaluate the effectiveness of the negative sampling strategy. It turns out that using 5 random negative samples for each claim performs the best.
For each claim with evidence, 10 to 15 sentences before and after the claims are chosen to be the evidence candidates.
The negative sampling strategy is also applied for the evidence candidates in the training set, where the ratio of positive samples (i.e., 7,278 pieces of evidence) to negative samples (i.e., 50,120 pieces of non-evidence) is 1:6.9. 
It turns out that using 1 random negative sample for each piece of evidence is the best. 

We implement the sentence-pair classification model and the multi-label model for CESC with the aid of SimpleTransformers \cite{rajapakse2019simpletransformers}.
The multi-task model for CEPE is based on the implementation of the multi-task framework by \citet{cheng2021argument}.
All models are run with V100 GPU.
We train our models for 10 epochs.
We experiment with two pre-trained models: BERT \cite{devlin2019bert} and RoBERTa \cite{liu2019roberta}.
Batch size is set as 128 for claim extraction and stance classification, and 16 for evidence extraction.
We use 1 encoding layer for the multi-task model, and other parameters are the same as the previous work.
\footnote{More details about hyper-parameter settings (i.e., batch sizes in the sentence-pair classification model, number of layers in the multi-task model), runtime and performance on the development set could be found in Appendix \ref{sec:appen_hyper}.}

For the claim and evidence extraction subtasks, besides Macro F$_1$ and Micro F$_1$, we also report the claim-class F$_1$ and the evidence-class F$_1$, respectively.
For CESC, we additionally report the claim-class F$_1$ of different stances (i.e., support and contest).
For the claim stance classification subtask, we report overall accuracy and F$_1$ for each class, as this task can be simply considered as a binary classification problem with balanced labels.
For CEPE, we report precision, recall, and F$_1$.

\subsection{Main Results on Existing Tasks}

\begin{figure}
\begin{tikzpicture}
\pgfplotsset{width=8cm,height=3.8cm,compat=newest}
\begin{axis}[
    xtick={1,2,3,4,5,6,7,8,9,10,11,12,13.3},
    ymin=33, ymax=55,
    xticklabels = {1,2,3,4,5,6,7,8,9,10,11,12,13.3},
    xticklabel style = {font=\fontsize{6}{1}\selectfont},
    yticklabel style = {font=\fontsize{6}{1}\selectfont},
    legend style={font=\fontsize{5}{1}\selectfont},
	ylabel={\footnotesize F$_1$ on claims},
	xlabel={\footnotesize \# negative samples for each claim},
	enlargelimits=0.1,
	legend style={at={(0.28,0.72)},anchor=south,legend columns=2}, 
	every axis plot/.append style={thick},
	tick label style={/pgf/number format/fixed},
    every node near coord/.append style={font=\tiny}
]

\addplot[blue] [mark=square*]  coordinates {
(1, 37.26) (2, 40.94) (3, 45.29) (4, 48.85)
(5, 50.35) (6, 48.84) (7, 47.07) (8, 47.99)
(9, 48.11) (10, 46.8) 
(11, 48.95) (12, 45.62)
(13.3, 34.76)
 };

\end{axis}
\end{tikzpicture}
\caption{Effect of negative sampling for claim extraction with RoBERTa-base model.}
\label{fig:claim_ns}
\end{figure}
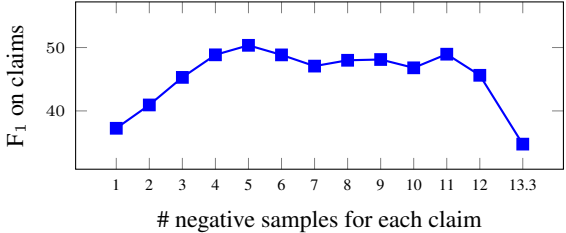

\paragraph{Claim Extraction Performance}
Table \ref{tab:claim} shows the performance on Task 1.
The classification model with pre-trained RoBERTa-base performs slightly better than with BERT-base-cased.
Recall that we adopt the negative sampling strategy for these two models by randomly selecting 5 negative samples during the training phase.
We also compare the performance of using different numbers of negative samples for each claim as shown in Figure \ref{fig:claim_ns}.
Generally speaking, the model performs better as the number of negative samples increases from 1 to 5, and starts to drop afterward.
As the ratio is more balanced, i.e., from no sampling (1:13.3) to 5 negative samples, the F$_1$ score increases as expected.
As the number of negative samples decreases further to 1, the ratio is even more balanced.
However, it sacrifices the number of training data, which leads to worse performance.

\begin{table}[t!]
	\centering
	    \resizebox{\columnwidth}{!}{
        \begin{tabular}{lccc}
        \toprule
        Models & Macro F$_1$ & Micro F$_1$ & Claim F$_1$\\
        \midrule
        BERT-base-cased & 72.08 & \textbf{92.51} & 48.08\\
        RoBERTa-base & \textbf{72.36} & 91.09 & \textbf{50.35}\\
        \bottomrule
        \end{tabular}}
    \caption{Claim extraction performance.
    }
	\label{tab:claim}
\end{table}

\begin{table}[t!]
	\centering
	    \resizebox{\columnwidth}{!}{
        \begin{tabular}{lccc}
        \toprule
        Models & Acc. & Support F$_1$ & Contest F$_1$\\
        \midrule
        BERT-base-cased  & 73.43 & 73.08 & 73.78 \\
        RoBERTa-base & \textbf{81.21} & \textbf{81.21} & \textbf{81.21} \\
        \bottomrule
        \end{tabular}}
    \caption{Stance classification performance.}
	\label{tab:stance}
\end{table}

\paragraph{Stance Classification Performance}
Table \ref{tab:stance} shows the performance on Task 2.
In both models, the F$_1$ scores on each stance are very close to each other, which is as expected because the two stances are balanced as shown in Table \ref{tab:data4}.
Although the pre-trained RoBERTa model outperforms the BERT model, there is still ample room for improvement as the accuracy of the RoBERTa model (81.21) is not relatively high for a binary classification task.
One possible reason is that some claim sentences are too long to intuitively show the stances.
For example, for the topic ``Should vaccination be mandatory'', a claim sentence ``Young children are often at increased risk for illness and death related to infectious diseases, and vaccine delays may leave them vulnerable at ages with a high risk of contracting several vaccine-preventable diseases.'' is classified as ``+1'' according to the human evaluation, but is predicted as ``-1'' from the RoBERTa model.

\begin{figure}[t!]
\begin{tikzpicture}
\pgfplotsset{width=8cm,height=3.8cm,compat=newest}
\begin{axis}[
    xtick={1,2,3,4,5,6.9},
    ymin=20, ymax=40,
    xticklabels = {1,2,3,4,5,6.9},
    xticklabel style = {font=\fontsize{6}{1}\selectfont},
    yticklabel style = {font=\fontsize{6}{1}\selectfont},
    legend style={font=\fontsize{5}{1}\selectfont},
	ylabel={\footnotesize F$_1$ on evidence},
	xlabel={\footnotesize \# negative samples for each piece of evidence},
	enlargelimits=0.1,
	legend style={at={(0.28,0.72)},anchor=south,legend columns=2}, 
	every axis plot/.append style={thick},
	tick label style={/pgf/number format/fixed},
    every node near coord/.append style={font=\tiny}
]

\addplot[blue] [mark=square*]  coordinates {
(1, 37.92) (2, 35.89) (3, 34.44) (4, 30.98)
(5, 30.18) (6.9, 20.34)
 };

\end{axis}
\end{tikzpicture}
\caption{Effect of negative sampling for evidence extraction with BERT-base-cased (C).}
\label{fig:evidence_ns}
\end{figure}
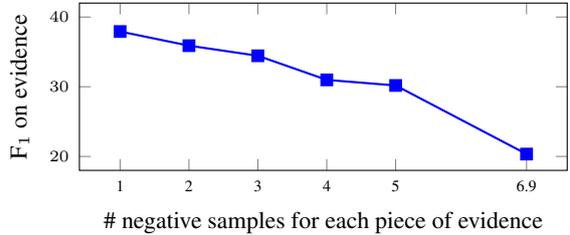

\paragraph{Evidence Extraction Performance}
Table \ref{tab:evidence} shows the performance on Task 3.
Again, the RoBERTa model performs better than the BERT model.
For this task, we experiment with two settings:
(1) given the topic and the claim (T+C), (2) only given the claim (C), to identify the evidence from the candidate sentences. 
For the (T+C) setting, we simply concatenate the topic and the claim as a sentence, and pair up with the evidence candidates to predict whether it is a piece of evidence of the given claim under the specific topic.
Comparing the results of these two settings, adding the topic sentences as inputs does not significantly improve the performance further, which suggests that claims have a closer relation with evidence, while the topic is not a decisive factor to evidence extraction.
Here, 1 negative sample for each evidence sentence is randomly selected.
The comparison of different numbers of negative samples is shown in Figure \ref{fig:evidence_ns}.
Unlike the trend shown in the claim extraction task, the model achieves the best performance when the ratio is exactly balanced at 1:1.

\begin{table}[t!]
	\centering
	    \resizebox{\columnwidth}{!}{
        \begin{tabular}{lccc}
        \toprule
        Models & Macro F$_1$ & Micro F$_1$ & Evi. F$_1$ \\
        \midrule
        BERT-base-cased (T+C) & 58.17 & 72.75 & 38.15 \\
        RoBERTa-base (T+C) & 62.43 & 78.13 & \textbf{40.89} \\
        \midrule
        BERT-base-cased (C) & 58.01 & 72.65 & 37.92 \\
        RoBERTa-base (C) & \textbf{63.37} & \textbf{80.29} & 40.16 \\
        \bottomrule
        \end{tabular}}
    \caption{Evidence extraction performance.
    }
	\label{tab:evidence}
\end{table}

\subsection{Main Results on Integrated Tasks}

For these two integrated tasks, we first use a pipeline method to pipe the best performing model on each corresponding subtask together, and then compare the overall performance with the proposed end-to-end models.

\begin{table}[t!]
	\centering
	    \resizebox{\columnwidth}{!}{
        \begin{tabular}{lcccc}
        \toprule
        Models & Macro F$_1$ & Micro F$_1$ & Support F$_1$ & Contest F$_1$\\
        \midrule
        Pipeline & 55.95 & 88.56 & 33.39 & 40.10 \\
        Multi-label & \textbf{60.25} & \textbf{91.22} & \textbf{38.34} & \textbf{47.31} \\
        \bottomrule
        \end{tabular}}
    \caption{CESC task performance.}
	\label{tab:cesc}
\end{table}

\paragraph{CESC Task Performance}
Table \ref{tab:cesc} shows the results of two approaches for the CESC task.
For both two methods, we randomly select 5 negative samples for each positive sample (i.e., claim) during training.
The pipeline model trains two subtasks independently and pipes them together to predict whether a sentence is a claim and its stance.
Although it achieves the best performance on each subtask, the overall performance is poorer than the multi-label model.
It shows that identifying the stances of the claims can benefit the claim extraction subtask, and such a multi-label model is beneficial to the integrated CESC task.

\begin{table}[t!]
	\centering
	    \setlength{\tabcolsep}{6mm}{
	    \resizebox{\columnwidth}{!}{
        \begin{tabular}{lccc}
        \toprule
        Models & Precision & Recall & F$_1$ \\
        \midrule
        Pipeline & 16.58 & 22.11 & 18.95 \\
        Traversal & 24.06 & \textbf{38.74} & 29.69 \\
        Multi-task & \textbf{43.54} & 30.57 & \textbf{35.92} \\
        \bottomrule
        \end{tabular}}
        }
    \caption{CEPE task performance.}
	\label{tab:cepe}
\end{table}

\paragraph{CEPE Task Performance}
Table \ref{tab:cepe} shows the overall performance comparison among different approaches.
Apart from the pipeline and the multi-task models as mentioned, we add another baseline model named ``traversal''.
In this model, all possible pairs of  ``topic + claim candidate''
and ``evidence candidate'' 
are concatenated and fed into the sentence-pair classification model.
Both the traversal model and the multi-task model outperform the pipeline model in terms of the overall F$_1$ score, which implies the importance of handling these two subtasks together.
The better performance of the multi-task model over the traversal model demonstrates the strong subtask coordination capability of the multi-task architecture. 

\begin{table}[t!]
	\centering
	\resizebox{\columnwidth}{!}{
        \begin{tabular}{p{7cm} ccc}
        \toprule
        \textbf{Topic: Should we ban human cloning} & \textbf{Gold} & \textbf{PL} & \textbf{MT} \\
        \midrule
        \textbf{Claim:} Cloning humans could reduce the impact of diseases in ways that vaccinations cannot. & \multirow{3}{*}{C} & \multirow{3}{*}{C} & \multirow{3}{*}{C} \\
        \midrule
        This method could help countries like Japan who are struggling with low birth rates. &  & \multirow{2}{*}{E} &  \\
        \cdashline{1-4}
        The Japanese culture could see a reduction of up to 40 million people by the year 2060 without the introduction of cloning measures. &  & \multirow{3}{*}{E} & \\
        \cdashline{1-4}
        Human cloning could help us to begin curing genetic diseases such as cystic fibrosis or thalassemia. & \multirow{3}{*}{E} & \multirow{3}{*}{E} & \multirow{3}{*}{E} \\
        \cdashline{1-4}
        Genetic modification could also help us deal with complicated maladies such as heart disease or schizophrenia. & \multirow{3}{*}{E} & & \multirow{3}{*}{E} \\
        \bottomrule
        \end{tabular}}
    \caption{Examples of model predictions for CEPE task.
    PL stands for the pipeline, and MT stands for the multi-task. 
    We select four sentences from the evidence candidates to demonstrate the prediction results here.}
	\label{tab:casestudy}
\end{table}

\subsection{Case Study}
We present a few examples in Table \ref{tab:casestudy} to compare the prediction results from the pipeline approach and the multi-task method for the CEPE task.
Given the topic ``should we ban human cloning'', both models successfully identify the claim sentence.
The first two sentences are not labeled as evidence supporting this claim based on the human annotation.
The multi-task model labels these two sentences correctly, while the pipeline model predicts them as evidence by mistake.
We notice that phrases of giving examples
(e.g., ``countries like'') and numbers (e.g., ``40 million'', ``year 2060'') are very common elements in evidence, which are the typical evidence types like demonstration with examples and numerical evidence.
We further explore the label predictions of these two sentences toward other claims and observe the pipeline approach classifies them as evidence as well.
Without understanding the true meaning of the sentences, the pipeline approach only learns the common words and the structure.
For the third evidence candidate, both models correctly predict this sentence and the extracted claim as a claim-evidence pair.
However, the pipeline model fails to identify the last evidence candidate sentence as a piece of evidence supporting the extracted claim.
This is plausibly because the claim and the last evidence candidate sentence share few vocabularies. 
Although ``genetic modification'' is different from ``cloning humans'', they still share some similarities in terms of semantic comprehension in the context, thus the second sentence can also support the claim.
Compared to the pipeline approach simply using the sentence-pair classification on the current sentences step by step, the multi-task model can learn a better sentence representation by utilizing the context information and coordinating two subtasks explicitly through the attention-guided multi-cross encoding layer, which finally leads to better performance.
See Appendix \ref{sec:more_eg} for more examples.



\section{Conclusions}
In this paper, we introduce a comprehensive and large dataset named IAM for argument mining to facilitate the study of multiple tasks involved in the debating system.
Apart from the existing primary argument mining tasks for debating, we propose two integrated tasks to work towards the debate automation, namely CESC and CEPE.
We experiment with a pipeline method and an end-to-end approach for both integrated tasks.
Experimental results and analysis are presented as baselines for future research, and demonstrate the value of our proposed tasks and dataset.
In the future, we will continue studying the relations among the argument mining subtasks and also explore more useful research tasks in the debating system.

\nocite{*}
\bibliography{anthology,custom}
\bibliographystyle{acl_natbib}

\clearpage
\appendix

\section{More Experimental Details}
\label{sec:appen_hyper}

\subsection{Hyper-parameters}

We manually tune the hyper-parameters in our models.
Table \ref{tab:claim_batchsize} shows the results on claim extraction task with different batch sizes from 8 to 128. Here, we use the pre-trained RoBERTa-base model. 5 negative samples are randomly chosen for each claim during training.
When the batch size is 128, the model achieves the best performance.

\begin{table}[H]
	\centering
    \resizebox{\columnwidth}{!}{
        \begin{tabular}{lcccc}
        \toprule
        Models & Batch size & Macro F$_1$ & Micro F$_1$ & Claim F$_1$\\
        \midrule
        RoBERTa-base & 8 & 68.44 & 91.24 & 41.65 \\
        RoBERTa-base & 16 & 70.92 & \textbf{92.17} & 45.94 \\
        RoBERTa-base & 32 & 71.59 & 90.97 & 48.82 \\
		RoBERTa-base & 64 & 72.11 & 91.89 & 48.85 \\
        RoBERTa-base & 128 & \textbf{72.36} & 91.09 & \textbf{50.35} \\
        \bottomrule
        \end{tabular}}
    \caption{Claim extraction performance with different batch sizes.
    }
	\label{tab:claim_batchsize}
\end{table}

Table \ref{tab:stance_batchsize} shows the results of using different batch sizes ranging from 8 to 128 for the stance classification task. Each model (i.e., BERT and RoBERTa) achieves the best performance when the batch size is 128.

\begin{table}[H]
	\centering
    \resizebox{\columnwidth}{!}{
	    \setlength{\tabcolsep}{6mm}{
        \begin{tabular}{lcc}
        \toprule
        Models & Batch size & Accuracy \\
        \midrule
        BERT-base-cased & 8 & 69.45 \\
        BERT-base-cased & 16 & 68.50 \\
        BERT-base-cased & 32 & 76.28 \\
        BERT-base-cased & 64 & 65.09 \\
        BERT-base-cased & 128 & 73.43 \\
        RoBERTa-base & 8 & 70.97 \\
        RoBERTa-base & 16 & 75.71 \\
        RoBERTa-base & 32 & 78.37 \\
        RoBERTa-base & 64 & 79.32 \\
        RoBERTa-base & 128 & \textbf{81.21} \\
        \bottomrule
        \end{tabular}}
        }
    \caption{Results of different batch sizes for stance classification task.}
	\label{tab:stance_batchsize}
\end{table}

Table \ref{tab:multitask_layers} shows the effect of using different numbers of layers in the multi-task model. More model details regarding each layer could be found in \cite{cheng2021argument}'s work. The multi-task model achieves the best F$_1$ score when the number of layers is 1.

\begin{table}[H]
	\centering
	    \setlength{\tabcolsep}{6mm}{
	    \resizebox{\columnwidth}{!}{
        \begin{tabular}{lccc}
        \toprule
        Layers & Precision & Recall & F$_1$ \\
        \midrule
        1 & 43.54 & \textbf{30.57} & \textbf{35.92} \\
        2 & 44.79 & 26.60 & 33.38 \\
        3 & \textbf{45.65} & 22.64 & 30.27 \\
        \bottomrule
        \end{tabular}}
        }
    \caption{Effect of different numbers of layers used in the multi-task model.}
	\label{tab:multitask_layers}
\end{table}

\subsection{Runtime and Validation Performance}

In Table \ref{tab:rt_dev}, we present the running time and the results on the development set of the multi-task model on the CEPE task.
As the number of layers increases, it requires a longer training time.

\begin{table}[H]
	\centering
	    \setlength{\tabcolsep}{3mm}{
	    \resizebox{\columnwidth}{!}{
        \begin{tabular}{lcccc}
        \toprule
        Layers & RT (min) & Dev P. & Dev R. & Dev F$_1$ \\
        \midrule
        1 & 15 & 33.31 & 20.66 & 25.50 \\
        2 & 22 & 39.81 & 18.68 & 25.43 \\
        3 & 29 & 40.59 & 18.36 & 25.28 \\
        \bottomrule
        \end{tabular}}
        }
    \caption{Runtime (RT) per epoch (minutes), the precision (P.), recall (R.) and F$_1$ on the development set of the multi-task model with differnt numbers of layers.}
	\label{tab:rt_dev}
\end{table}

\section{More Case Study}
\label{sec:more_eg}

Table \ref{tab:more_eg} shows more example predictions generated by the pipeline approach and the multi-task model for the CEPE task.
In these examples, the multi-task model identifies most of the claim-evidence pairs while the pipeline method fails to do so.
For the second topic which is shown earlier in Section \ref{sec:data_annotation}, the pipeline model fails to detect the claim sentence nor the evidence sentence.

\begin{table}[ht]
	\centering
	\resizebox{\columnwidth}{!}{
        \begin{tabular}{p{7cm} ccc}
        \toprule
        \textbf{Topic: Should we fight for the Olympics} & \textbf{Gold} & \textbf{PL} & \textbf{MT} \\
        \midrule
        \textbf{Claim:} These often impose costs for years to come. & \multirow{2}{*}{C} & \multirow{2}{*}{C} & \multirow{2}{*}{C} \\
        \midrule
        Sydney’s Olympic stadium costs the city \$30 million a year to maintain. & \multirow{2}{*}{E} & & \multirow{2}{*}{E} \\
        \cdashline{1-4}
        Beijing’s famous “Bird’s Nest” stadium cost \$460 million to build and requires \$10 million a year to maintain, and sits mostly unused. & \multirow{3}{*}{E} & & \multirow{3}{*}{E} \\
        \midrule \bottomrule \hline
        
        \textbf{Topic: Will artificial intelligence replace humans} & & & \\
        \midrule
        \textbf{Claim:} Any job that involves repetitive tasks is at risk of being replaced. & \multirow{2}{*}{C} & & \multirow{2}{*}{C} \\
        \midrule
        In 2017, Gartner predicted 500,000 jobs would be created because of AI, but also predicted that up to 900,000 jobs could be lost because of it. & \multirow{3}{*}{E} & & \multirow{3}{*}{E} \\
        \midrule \bottomrule \hline
        
        \textbf{Topic: Should we implement the network real-name system} & & & \\
        \midrule
        \textbf{Claim:} Real-name policy blurs the boundaries between personal information and personal privacy. & \multirow{2}{*}{C} & \multirow{2}{*}{C} & \multirow{2}{*}{C} \\
        \midrule
        Due to the vague boundaries between privacy and personal information, today people are willing to distinguish this boundary between online behavior and offline ID. & \multirow{3}{*}{E} & & \\
        \cdashline{1-4}
        For example, as an Internet user, my words and deeds on the Internet, personal information published, such as political positions, belong to my personal information. & \multirow{3}{*}{E} & & \multirow{3}{*}{E} \\
        \cdashline{1-4}
        But once it matches my true identity, it is personal privacy. & \multirow{3}{*}{E} & & \multirow{3}{*}{E} \\
        \bottomrule

        \end{tabular}}
    \vspace{-2mm}
    \caption{More example predictions of the CEPE task.}
	\label{tab:more_eg}
	\vspace{-3mm}
\end{table}

\end{document}